\documentclass[10pt,twocolumn,letterpaper]{article}

\usepackage[pagenumbers]{iccv} 
\usepackage{diagbox}
\definecolor{iccvblue}{rgb}{0.21,0.49,0.74}
\usepackage[pagebackref,breaklinks,colorlinks,allcolors=iccvblue]{hyperref}
\usepackage[accsupp]{axessibility}  

\def\paperID{7241} 
\def\confName{ICCV}
\def\confYear{2025}

\definecolor{cvprblue}{rgb}{0.21,0.49,0.74}
\usepackage{marvosym}
\usepackage{float}
\usepackage{multirow}
\usepackage{diagbox}

\newcommand{\best}[1]{\textcolor{red}{#1}}
\newcommand{\second}[1]{\textcolor{blue}{#1}}

\newcommand{\szm}[1]{\textcolor{black}{#1}}

\newcommand{\yr}[1]{\textcolor{black}{#1}}
\newcommand{\tyz}[1]{\textcolor{black}{#1}}
\newcommand{\yrnew}[1]{\textcolor{black}{#1}}

\title{Pinco: Position-induced Consistent Adapter for Diffusion Transformer\\ in Foreground-conditioned Inpainting}

\author{
Guangben Lu$^{1,2,\ast}$ ~~
Yuzhen Du$^{\yrnew{1,}2,\ast}$ ~~
Yizhe Tang$^{1}$ ~~
Zhimin Sun$^{1,2}$ ~~
Ran Yi$^{1}$\textsuperscript{\Letter} ~~\\
Yifan Qi$^{2}$ ~~
Tianyi Wang$^{2}$ ~~
Lizhuang Ma$^{1}$ ~~
Fangyuan Zou$^{2}$ \\ 
\normalsize  $^1$Shanghai Jiao Tong University \quad $^2$Tencent
\\ \normalsize \{tangyizhe, ranyi\textsuperscript{\Letter}\}@sjtu.edu.cn \quad ma-lz@cs.sjtu.edu.cn 
\\ \normalsize \{lucasgblu, yuzhendu, threecatsun, ivanqi, joshtywang, ericfyzou\}@tencent.com
\\ \normalsize $\ast$ Equal contribution ~~~{\Letter} Corresponding author
}

\begin{document}

\twocolumn[{%
\maketitle
\begin{figure}[H]
\vspace{-0.4in}
\hsize=\textwidth 
\centering
\includegraphics[width=0.85\textwidth]{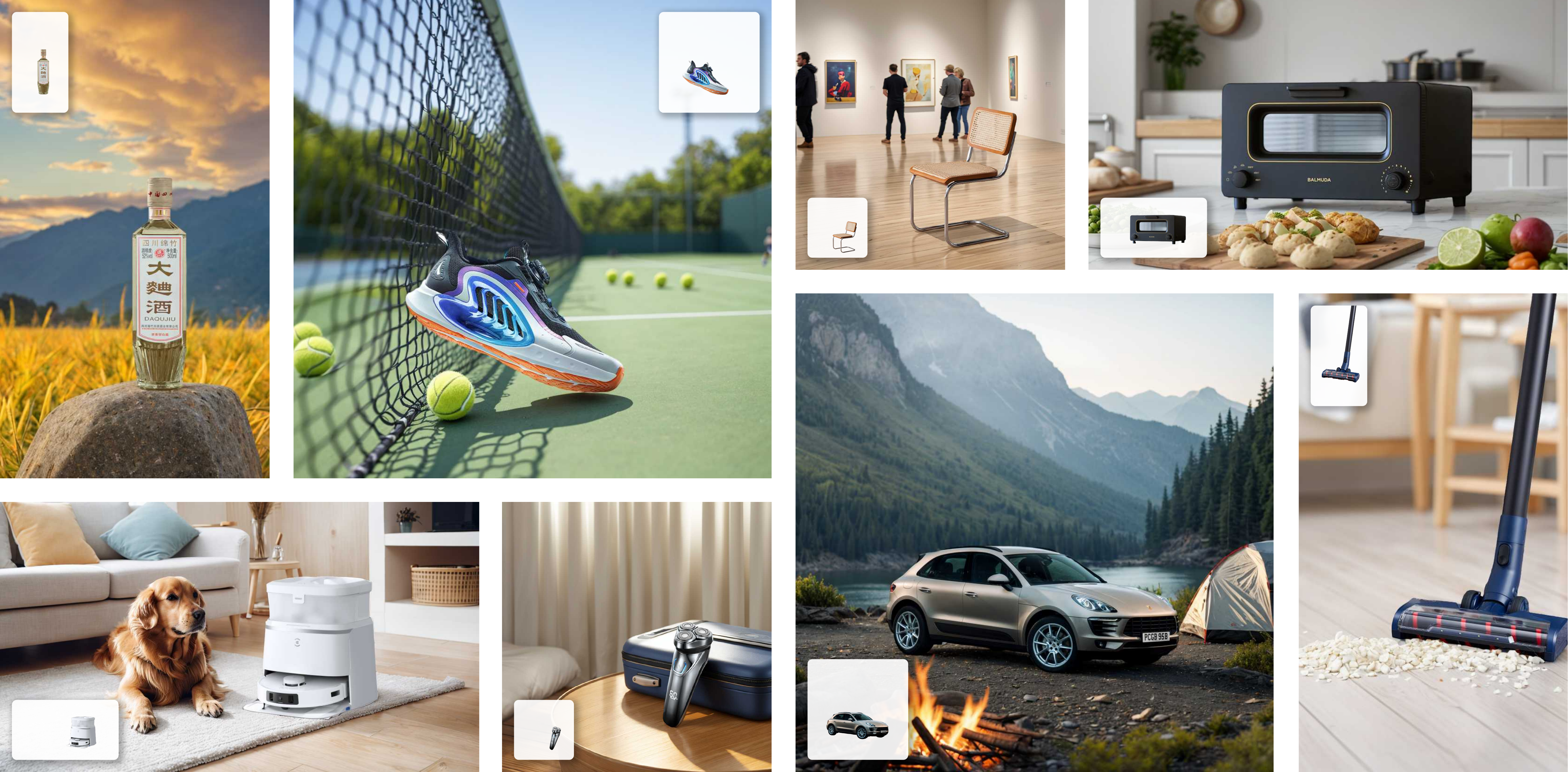}
\caption{
Pinco generates high-quality images with rich and diverse backgrounds from given foreground subjects and text descriptions. The four images on the left are generated by Hunyuan-Pinco and those on the right by Flux-Pinco, exhibiting its outstanding capabilities in foreground consistency preservation, rational spatial arrangements, and diverse background generation.}
\end{figure}
}]
\vspace{-0.4in}

\maketitle


\begin{abstract}
Foreground-conditioned inpainting aims to seamlessly fill the background region of an image by utilizing the provided foreground subject and a text description. While existing T2I-based image inpainting methods can be applied to this task, they suffer from issues of subject shape expansion,
distortion, or impaired ability to align with the text description, resulting in inconsistencies between the visual elements and the text description. To address these challenges, we propose Pinco, a plug-and-play foreground-conditioned inpainting adapter that generates high-quality backgrounds with good text alignment while effectively preserving the shape of the foreground subject. Firstly, we design a Self-Consistent Adapter that integrates the foreground subject features into the layout-related self-attention layer, which helps to alleviate conflicts between the text and subject features by ensuring that the model can effectively consider the foreground subject's characteristics while processing the overall image layout. Secondly, we design a Decoupled Image Feature Extraction method that employs distinct architectures to extract semantic and spatial features separately, significantly improving {subject feature extraction} and ensuring high-quality preservation of the subject's shape. Thirdly, to ensure precise utilization of the extracted features and to focus attention on the subject region, we introduce a Shared Positional Embedding Anchor, greatly improving the model's understanding of subject features and boosting training efficiency. Extensive experiments demonstrate that our method achieves superior performance and efficiency in foreground-conditioned inpainting. 
\end{abstract}    
\section{Introduction}
\label{sec: Intro}

{Visual generative models~\cite{ldm, chen2023pixart,hu2025improving} have made significant progress recently }
{and} have demonstrated exceptional potential in various applications, including image editing~\cite{huang2024diffusion, li2023virtual, yi2024feditnet++, ma2024taming}, image restoration~\cite{li2017generative, Audio_driven, lugmayr2022repaint,tian2023pyramid,du2024ld,du2024exploring}, and generative safety~\cite{wang2023noise, liu2025learning, sun2023contrastive, sun2024rethinking}. Among these applications, a notably promising yet challenging task is \textbf{foreground-conditioned inpainting}, which focuses on generating high-quality backgrounds based on a provided text description while maintaining the integrity of the given foreground subject. It is closely related to the conventional text-guided image inpainting task ~\cite{ldm, chen2023pixart, zhuang2023powerpaint, ju2024brushnet}, which involves inpainting content within a small region based on a brief text description. However, the foreground-conditioned inpainting task presents greater challenges due to the following reasons: 1) It requires inpainting a larger area (the background region) with a more complex text description, and 2) It is crucial to preserve the integrity of the foreground subject's shape while ensuring harmonious coordination between the foreground subject and the background. 

\begin{figure}[t]
    \centering
    \includegraphics[width=0.46\textwidth]{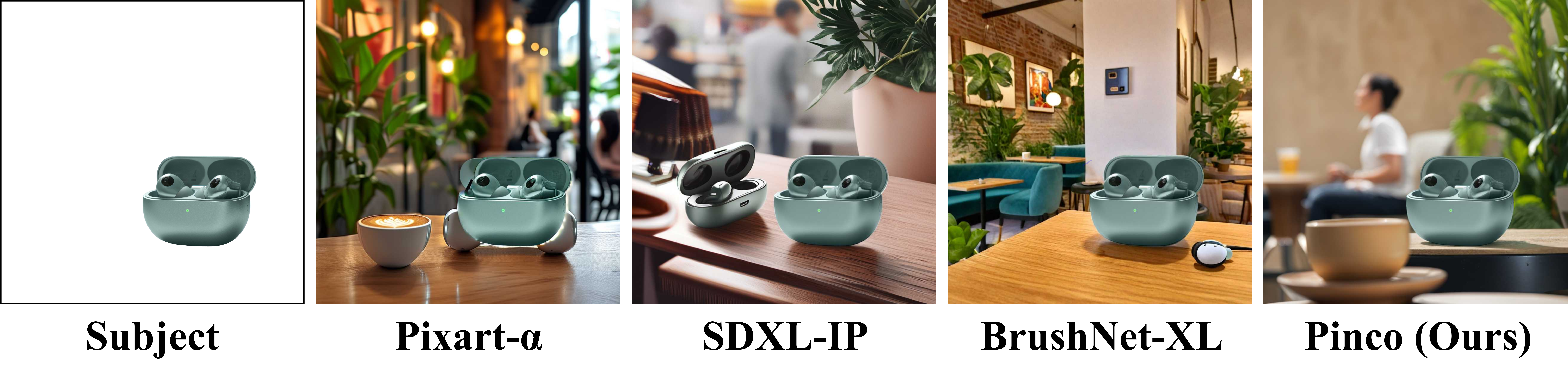}
    \caption{
    Foreground-conditioned inpainting results of three existing T2I Inpainting methods,
    Pixart-$\alpha$ (sampling strategy modification), SDXL-Inpainting (Model fine-tuning), BrushNet (Conditional injection), and ours. Existing methods suffer from issues such as shape expansion and unreasonable spatial relationships between the foreground subject and background.}
    \label{fig: methods comp}
    \vspace{-0.2in}
\end{figure}

{Conventional text-guided image inpainting methods based on T2I diffusion models can be classified into three categories: 1) Sampling modification methods~\cite{avrahami2022blended,corneanu2024latentpaint,zhang2023adding, chen2023pixart, avrahami2023blended, manukyan2023hd}, which adjust the diffusion sampling strategy to achieve inpainting, such as through latent space replacement. Although these methods can be seamlessly adapted to various diffusion backbones, they may disrupt the original latent distribution, leading to irrational layout compositions or unreasonable object placements (Fig.~\ref{fig: methods comp}). 2) Model fine-tuning methods ~\cite{sdxl, zhuang2023powerpaint, yang2023magicremover,hu2024sara}, which typically involve modifying the network structure of pre-trained T2I models and utilizing relatively small-scale datasets. These methods can cause the model to forget pre-trained knowledge and make it prone to fitting the distribution of the fine-tuned training dataset, ultimately affecting the quality and diversity of the generated images. 3) Conditional injection methods ~\cite{controlnet, ju2024brushnet, he2024freeedit, firefly2023, alimama_FLUX}, which incorporate information from the preserved region into the frozen T2I model via a side-branch mechanism. Most of these methods ~\cite{ju2024brushnet, firefly2023, he2024freeedit} employ the ControlNet ~\cite{controlnet} structure, replicating half or the entire T2I model for inpainting injection. However, as the scale of the T2I model increases, these side-branch models become increasingly cumbersome and impractical to implement. Additionally, adding extra control information after the completion of text cross-attention may result in a decline in text-image consistency.}

{To address these issues,} we propose \textbf{Pinco}, {a powerful yet efficient} plug-and-play {adapter that empowers DiT-based models for {consistent} foreground-conditioned inpainting}. We design three innovative modules: 
1) \textbf{{Self-Consistent Adapter}}. 
To effectively inject the subject features into the base T2I model, we need a suitable condition injection mechanism, where 
conventional condition injection methods based on image prompt adapter inject the
feature through cross-attention layer and integrate it with the text cross-attention in the base model. 
However, such integration by directly adding outputs of subject cross-attention and text cross-attention can easily cause conflicts between text and the injected subject information, resulting in a mismatch between the text and generated background. Therefore, we propose a Self-Consistent Adapter, which makes an innovative shift by integrating subject-aware attention into the self-attention layer, effectively alleviating the conflicts between text and subject features. 
2) \textbf{Decoupled Image Feature Extraction}. The conventional feature extractors, CLIP and VAE encoder, used in controllable T2I models are both suboptimal for our subject feature extraction: CLIP encoder only captures abstract global semantic information, while VAE encoder provides a limited amount of shape features, which is insufficient to meet the requirements of strict contour preservation in our task. To address these issues, we propose a novel Semantic-Shape Decoupled Extractor, using different architectures for the extraction of semantic and shape features from different sources, ensuring effective extraction of subject features with fine details of subject shape, achieving high-quality preservation of the subject outline and effectively avoiding expansion. 
3) \textbf{Shared Positional Embedding Anchor}. To ensure precise utilization of subject features, subject-aware attention should be concentrated in the subject region. In light of this, we propose a Shared Positional Embedding Anchor that combines positional embeddings with subject features before the calculation of subject-aware attention. This approach achieves a decay in attention weights as the distance from the subject region increases, thereby concentrating attention inside the subject region. This operation greatly enhances the model's understanding of subject features and improves training efficiency. Extensive experiments demonstrate that our method achieves superior foreground-conditioned inpainting performance and efficiency.

We summarize our contributions in four-fold:

\begin{enumerate}
    \item We propose Pinco, a powerful yet efficient plug-and-play adapter that can be seamlessly integrated with DiT-based models to achieve high-quality foreground-conditioned inpainting. With a small number of training parameters, it only increases negligible inference latency compared to the base model, while ensuring high-quality generation of rich and diverse backgrounds.
    \item We propose Self-Consistent Adapter, a mechanism that injects subject feature through cross-attention, and innovatively integrates this subject-aware attention into the self-attention layer of the base model, {effectively alleviating the conflicts between text and subject features.}
    \item We propose a Decoupled Image Feature Extraction {strategy}, which decouples the extraction of {the} semantic and spatial features {of the subject, ensuring effective extraction of subject features with fine details of subject shape,} and enhancing the preservation of the subject shape.
    \item We propose a Shared Positional Embedding Anchor, which combines positional embedding with subject features before attention calculation to constrain the activated region in the subject-aware attention, ensuring precise utilization of subject features.
\end{enumerate}

\section{Related Work}
\label{sec: Rel}

\subsection{Controllable Image Generation}
There are various approaches to {control} generative diffusion models, some of which may represent promising directions for subject-driven, prompt-guided image synthesis. Certain methods~\cite{ruiz2023dreambooth, chen2023disenbooth,mokady2023null,gal2022image} {achieve personalization by individually fine-tuning for each given subject.} {Despite their success, these methods require re-tuning to each new concept and fail to maintain the exact details of the subject}. To address these challenges, some methods~\cite{chen2024anydoor,elarabawy2022direct,ye2023ip,controlnet,mou2024t2i,hu2025hunyuancustom} {reuse certain modules of the base model to achieve general feature extraction from reference images,  and inject these extracted features into the frozen base model without subject-specific tuning.} {However, as the scale of the T2I models increases, these methods become increasingly impractical.} In contrast, IP-Adapter~\cite{ye2023ip} utilizes a lightweight decoupled attention mechanism to incorporate {semantic} features {of the reference subject} into the model. Another methods~\cite{chefer2023attend,liu2023cones,liu2024cones,hertz2022prompt} {directly alter the attention map during sampling to facilitate subject generation, which carries the risk of compromising the inherent textual coherence. }

\subsection{Image Inpainting for Foreground Condition}
With the impressive performance of T2I diffusion models~\cite{ldm, sdxl, dit, sd3} in image generation, a variety of {diffusion-based }inpainting methods~\cite{ldm, chen2023pixart, zhuang2023powerpaint, ju2024brushnet, tang2025ata, eshratifar2024salient, flux-fill} have been proposed. 1) some methods~\cite{avrahami2022blended,avrahami2023blended,corneanu2024latentpaint,zhang2023adding,chen2023pixart} utilize a classic unmask region copying strategy, replacing known areas with the original image during the diffusion model's inference. However, this approach is rigid and often overlooks the semantic information of the foreground subject, causing shape expansion problems. 2) To avoid this problem, some methods~\cite{manukyan2023hd,yang2023magicremover, zhuang2023powerpaint} ensure the preservation of the subject by modifying and fine-tuning the model. PowerPaint~\cite{zhuang2023powerpaint} fine-tunes a text-to-image model with two learnable task prompts to achieve the subject semantic understanding. However, this approach can be challenging to train and may compromise the model's inherent text-to-image capabilities, leading to unalignment between the generated background and the input text. 3) Other methods~\cite{controlnet, ju2024brushnet, he2024freeedit, alimama_FLUX} inject the information of known area by side-branch injection. BrushNet~\cite{ju2024brushnet} adopts a hierarchical approach by gradually incorporating the full UNet feature layer-by-layer into the pre-trained UNet.  {These ControlNet-style methods have extensive training parameters, and the pattern of injecting features after text cross-attention can lead to a deterioration of text alignment, as the visual features may dominate the generation.}
{To address these challenges}, we propose Pinco, a plug-and-play and lightweight adapter method, which requires very few parameters while efficiently injecting the characteristics of the subject.

\section{Preliminaries}
\label{sec: Pre}
\subsection{Diffusion Models}
The diffusion models~\cite{rombach2021highresolution, Tianyi1} encompass forward and reverse processes. In the forward process, a noise is added to a clean image $ x_0 $ in a closed form. In the reverse process, given input noise $ x_T $ sampled from a random Gaussian distribution, a trainable network $ \epsilon_{\theta} $ estimates the noise at each step $ t $ conditioned on $ c $. The diffusion model training aims to optimize the denoising network $\epsilon_{\theta}$ so that it can accurately predict the noise given the condition $ c $:
\begin{equation}
    \mathcal{L} =\mathbb{E}_{x_0, c, \epsilon\sim\mathcal{N}(0,1), t} \left[ \| \epsilon - \epsilon_{\theta}(x_t, t, c) \|^2_2 \right].
\end{equation}
{In our scenario, the condition $c$ includes detailed text descriptions as well as independent foreground regions.}

\begin{figure*}[t]
    \centering
    \includegraphics[width=0.95\textwidth]{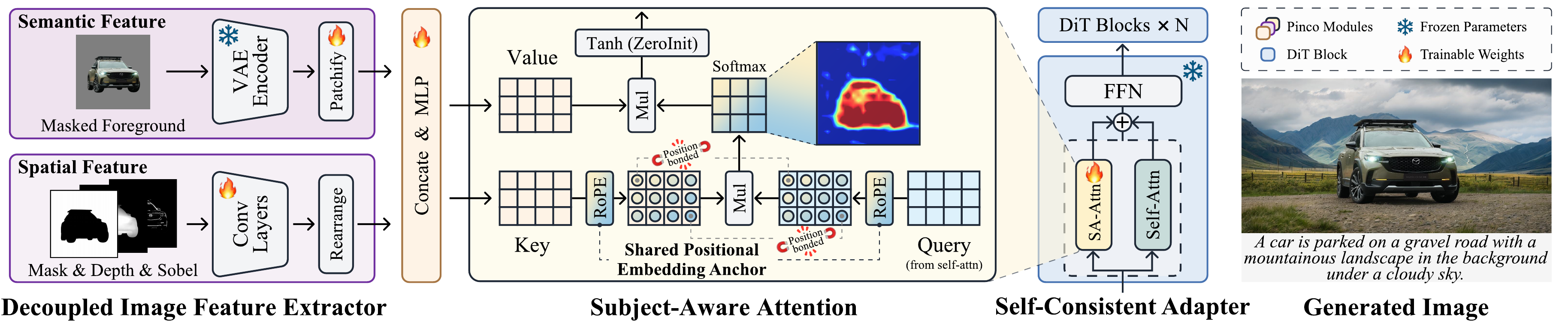}
    \caption{The framework of our \textbf{Pinco}, 
    a plug-and-play inpainting adapter that can be seamlessly integrated with a text-to-image DiT model for consistent foreground-conditioned inpainting. Pinco consists of three modules: a \textbf{Decoupled Feature Extractor} used to extract subject feature, a \textbf{Shared Positional Embedding Anchor} used to ensure foreground attention, and a \textbf{Self-Consistent Adapter} injecting subject feature on the self-attention layer.}
    \label{fig:model_archi}
    \vspace{-0.2in}
\end{figure*}

\subsection{Diffusion Transformer (DiT)} 

Pinco is built upon the \szm{Diffusion Transformer (DiT)~\cite{peebles2023scalable}, which is} a transformer-based diffusion method that operates on latent patches with a series of transformer blocks (DiT Blocks). Notably, the DiT Block enhances standard normalization layers with adaptive layer normalization (AdaLN).
\szm{Additionally, the DiT Block incorporates Rotary Positional Embedding (RoPE)~\cite{su2024roformer}, which captures both absolute and relative positional dependencies and introduces a two-dimensional RoPE to extend its functionality to the image domain.}
\yr{In this work, Pinco is integrated with two kinds of DiT models: 1)}
Hunyuan-DiT~\cite{li2024hunyuan}\yr{, which} further utilizes cross-attention mechanism for fine-grained text understanding similar to Stable Diffusion~\cite{ldm};
\yr{2)} \szm{FLUX.1~\cite{flux2024}, \yr{a MM-DiT model that} emphasizes the use of MM Attention~\cite{esser2024scalingrectifiedflowtransformers} to enhance the alignment between text and visual information.}

\section{Method}
\label{sec: Method}

\textbf{Overview.} Pinco is a plug-and-play inpainting adapter that can be seamlessly integrated with a text-to-image DiT model to enable consistent foreground-conditioned inpainting. The inputs to our inpainting system consist of a subject image $I$, a mask image $m$, and a text prompt $T$ (describing the desired background), with the subject depth image $d$ and Sobel image $s$ as conditioning signals. Pinco aims to inpaint the background region so that the generated background is consistent with the text description and visually natural, while the foreground subject remains unchanged.

Fig.~\ref{fig:model_archi} shows the framework of Pinco and base DiT model.
Pinco first extracts the semantic and spatial features of the subject image through a \textbf{Decoupled Image Feature Extractor} (Sec.~\ref{sec: Image Encoder}). Then Pinco injects the extracted subject feature into the base model through a \textbf{Self-Consistent Adapter} (Sec.~\ref{sec: Adapter}), which designs a subject-aware attention to inject subject features, and innovatively integrates it into the self-attention layer of the base model. Before the subject-aware attention calculation, Pinco designs a \textbf{Shared Positional Embedding Anchor} (Sec.~\ref{sec: Rope}), which combines positional embedding with the subject feature to constrain the activated area in subject-aware attention to the subject region. After the subject information is effectively injected and utilized, we employ the diffusion denoising of the base model to generate high-quality backgrounds.

\begin{figure*}[t]
    \centering
    \includegraphics[width=0.98\textwidth]{sec/assets/Vision_Result_new_new.pdf}
    \caption{Quantitative comparison of state-of-the-art inpainting methods and our \textit{Pinco}. Our method is capable of generating coherent, detailed and rational backgrounds following the provided text, while effectively mitigating the problem of subject expansion.}
    \label{fig: vision comp}
    \vspace{-0.2in}
\end{figure*}

\subsection{Self-Consistent Adapter}~\label{sec: Adapter}
In foreground-conditioned inpainting, the features of the foreground subject need to be injected into the base T2I diffusion model.
Classical methods for adding image conditions in T2I models mainly fall into three categories: 1) Concatenating condition images through input channels, which requires significant training overhead for fine-tuning. 2) Creating a side branch injection by copying part of the base model~\cite{controlnet, ju2024brushnet}, which adds many parameters and can slow down inference. 3) Training a lightweight image prompt adapter~\cite{ye2023ip, wang2024instantid,liu2024towards, tumanyan2023plug} through decoupled cross-attention mechanism, which has been proven effective. 
A na\"ive way to use such adpater is designing a cross-attention between the latent and subject features, and integrating it with the text cross-attention output.However, using this approach can lead to a disharmonious result. Since the text is often complex and primarily about the background, such integration can cause conflicts between text features and subject features, resulting in: 1) subject outward expansion or distortion in the generated image; and 2) a mismatch between the text and the generated background.

To tackle these challenges, we draw inspiration from the analyses of cross-attention and self-attention presented in previous works~\cite{liu2024towards,tumanyan2023plug}, which show that the self-attention map can be effectively leveraged to preserve the spatial structural characteristics of the original image. This observation indicates that self-attention is more closely aligned with the layout of the final output, thereby presenting a promising pathway for enhancing the results of foreground-conditioned inpainting. Building on this understanding, we propose \textbf{Self-Consistent Adapter}, which makes an innovative shift by integrating subject-aware attention directly into the self-attention layer.  For DiT with normal attention~\cite{dit, li2024hunyuan}, 
subject-aware attention is the cross-attention between the latent and subject features,
and our mechanism can be expressed as follows:
\begin{equation}
\setlength\abovedisplayskip{6pt}
\setlength\belowdisplayskip{6pt}
\begin{aligned}
    Z = \alpha & \odot \text{Self-Attention}(Q, K, V) \\
     + \beta & \odot \text{Cross-Attention}(Q, K_{sub}, V_{sub}),
\end{aligned}
\label{eq:saa}
\end{equation}

where $\alpha$ and $\beta$ are parameters used to tune the injection strength, $Q$ is the query, $K, K_{sub}, V$ and $V_{sub}$ are the key and value calculated from latents and subject feature. We also employ Zero-init Tanh Gating~\cite{zhang2023llama} to progressively control the strength of injection into the base model.
This approach avoids conflicts between subject features and text, thereby better preserving the subject shape in the generated image and maintaining the alignment between the background and text.

We find our Self-Consistent Adapter is also applicable to MM-DiT~\cite{esser2024scalingrectifiedflowtransformers, flux2024} that uses MM-Attention, a variant of self-attention. For MM-DiT, we design subject-aware attention as a MM-Attention that fuses the latent and subject features (detailed architecture in \#Suppl Fig.~S1). 
The modified attention mechanism is as follows:

\begin{equation}
\setlength\abovedisplayskip{6pt}
\setlength\belowdisplayskip{6pt}
\begin{aligned}
    Z = \alpha & \odot \text{MM-Attention}(Q, K, V) \\
    + \beta & \odot \text{MM-Attention}(Q, [K, K_{sub}], [V, V_{sub}]),
\end{aligned}
\label{eq: mm-attention}
\end{equation}
where
$[\cdot, \cdot]$ denotes concatenation across tokens dimension.

\subsection{Decoupled Image Feature Extractor}\label{sec: Image Encoder}

For T2I-model-based foreground-conditioned inpainting, the features of the foreground subject need to be extracted and injected into the base model, where the subject feature is commonly extracted from the concatenation of the subject image, the mask image and other images related to the subject (\textit{e.g.}, depth map and Sobel image).
Previously, image feature extraction of controllable T2I models~\cite{ye2023ip,sdxl, zhuang2023powerpaint, ju2024brushnet} mainly relied on CLIP or VAE encoder. 
However, the CLIP image encoder primarily captures the high-level global semantic information of the overall image, while the VAE encoder can hardly preserve the strict contour of the subject, both of which cannot meet the requirement of foreground high-consistency.
Therefore, we propose a Semantic-Shape Decoupled Image Feature Extractor to ensure the sufficient extraction of both semantic and shape information of the given subject.

Our Semantic-Shape Decoupled feature extractor consists of three parts: semantic feature extraction, shape feature extraction, and feature fusion. 
\textbf{1) Semantic Feature Extraction.} 
The semantic and textural features of the subject are mainly extracted from the 3-channel (RGB) subject image (with background removed). 
To align the extracted feature with the data distribution of the pre-trained diffusion model, we reuse the original VAE encoder $\varepsilon$ as the semantic feature extractor, which reduces the overhead of additional training. 
\textbf{2) Shape Feature Extraction.} 
It is crucial to maintain the details of the subject's shape in foreground-conditioned inpainting. 
Inaccurate shape feature extraction can lead to the expanding of the subject shape in the output image. 
Therefore, to accurately extract subject shape features, we leverage the subject's mask, depth map, and Sobel images to supplement the shape details of the subject. To ensure effective extraction of the local shape details, we construct a convolutional feature extractor to extract shape features. 
\textbf{3) Feature Fusion.}
After extracting semantic and shape features, both features are channel-wise concatenated together, and an MLP module is used to fuse them:

\begin{equation}
\setlength\abovedisplayskip{6pt}
\setlength\belowdisplayskip{6pt}
\begin{aligned}
    F = \text{MLP}([\varepsilon(I), \text{Conv}([m, d, s])]).
\end{aligned}
\end{equation}

Through this decoupled feature extraction method, we can not only fully utilize the information contained in the provided condition images, but also extract fine details of the subject shape, thereby significantly reducing the outward expansion of the subject shape in the output image.

\begin{table*}[t]
\centering
\caption{Quantitative comparison between Pinco and previous methods. Pinco demonstrates the best subject shape preservation and text alignment effects, while also having very few training parameters.}
\label{tab: Metric Comp}
\vspace{-0.05in}
\setlength\tabcolsep{5pt}
\scalebox{0.63}{
\begin{tabular}{cc|c|c|c|ccc|cc|c}
\hline
\multicolumn{2}{c|}{\multirow{2}{*}{\textbf{Architecture \& Backbone}}}                                 & \multirow{2}{*}{\diagbox{\textbf{Methods}}{\textbf{Metrics}}}           & \textbf{ Trainable Parameters} & \textbf{Image Quality} & \multicolumn{3}{c|}{\textbf{Foreground Consistency}}                 & \multicolumn{2}{c|}{\textbf{Text Alignment}} & \textbf{Rationality} \\ \cline{4-11} 
\multicolumn{2}{c|}{}        &            & \textbf{TPR}              & \textbf{FID $\downarrow$}         & \textbf{OER(SAM2.1)$\downarrow$} & \textbf{OER(BiRefNet) $\downarrow$} & \textbf{LPIPS $\downarrow$}  & \textbf{VQAScore$\uparrow$}   & \textbf{FV2Score $\downarrow$}   & \textbf{GPT4o$\uparrow$}      \\ \hline
\multicolumn{1}{c|}{\multirow{8}{*}{\textbf{UNet}}} & \multirow{4}{*}{\textbf{SD1.5}}       & \textbf{SD1.5 ControlNet}  & 29.59$\%$                   & 84.28                  & 22.00$\%$               & 52.28$\%$                  & 0.008555          & 0.849                & 22.83$\%$               & 4.012                \\
\multicolumn{1}{c|}{}                               &                                       & \textbf{HD-Painter}        & 0.00$\%$                    & 88.75                  & 22.04$\%$              & 37.56$\%$                  & 0.007171          & 0.813                & 13.26$\%$               & 4.624                \\
\multicolumn{1}{c|}{}                               &                                       & \textbf{PowerPaint-V2}     & 0.06$\%$                    & 88.52                  & 33.80$\%$               & 44.89$\%$                  & 0.010514          & 0.801                & 22.74$\%$               & 4.669                \\
\multicolumn{1}{c|}{}                               &                                       & \textbf{BrushNet-SD1.5}    & 41.86$\%$                   & \best{\textbf{84.16}}         & 19.60$\%$               & 36.19$\%$                  & 0.005779          & 0.868                & 19.67$\%$               & 4.335                \\ \cline{2-11} 
\multicolumn{1}{c|}{}                               & \multirow{4}{*}{\textbf{SDXL}}        & \textbf{SDXL inpainting}   & 100.00$\%$                  & 83.60                  & 25.24$\%$               & 30.77$\%$                  & 0.007946          & 0.847                & 13.96$\%$               & 4.536                \\
\multicolumn{1}{c|}{}                               &                                       & \textbf{LayerDiffusion}    & 24.67$\%$                   & 107.98                 & 20.31$\%$               & 27.81$\%$                  & 0.007996          & 0.864                & 17.34$\%$               & 4.358                \\
\multicolumn{1}{c|}{}                               &                                       & \textbf{BrushNet-SDXL}     & 12.69$\%$                   & 87.00                  & 28.25$\%$               & 18.91$\%$                  & 0.005776          & 0.870                & 9.83$\%$                & 4.621                \\
\multicolumn{1}{c|}{}                               &                                       & \textbf{Kolors Inpainting} & 100.00$\%$                  & 85.59                  & 21.47$\%$               & 14.40$\%$                  & \best{\textbf{0.004223}} & 0.891                & 5.27$\%$                & 4.418                \\ \hline
\multicolumn{1}{c|}{\multirow{4}{*}{\textbf{DiT}}}  & \multirow{2}{*}{\textbf{Hunyuan-DiT}} & \textbf{HY-ConntrolNet}    & 31.20$\%$                   & 85.31                  & 11.78$\%$               & 11.95$\%$                  & 0.006198        & 0.877                & 5.00$\%$                & 4.576                \\
\multicolumn{1}{c|}{}                               &                                       & \textbf{HY-Pinco (ours)}          & 11.37$\%$                   & \second{\textbf{84.25}}         & \second{\textbf{11.51$\%$}}      & \second{\textbf{10.00$\%$}}         & 0.004441 & 0.901       & \second{\textbf{3.16$\%$}}       & \best{\textbf{4.790}}       \\ \cline{2-11} 
\multicolumn{1}{c|}{}                               & \multirow{2}{*}{\textbf{Flux}}        & \textbf{Flux Controlnet}   & 15.25$\%$                   & 111.91                 & 28.11$\%$               & 22.30$\%$                  & 0.004502          & \second{\textbf{0.916}}       & 13.17$\%$               & 4.629                \\
\multicolumn{1}{c|}{}                               &                                       & \textbf{Flux-Pinco (ours)}        & 12.56$\%$                   & 103.41                 & \best{\textbf{7.87$\%$}}       & \best{\textbf{6.84$\%$}}          &   \second{\textbf{0.004239}}       & \best{\textbf{0.918}}       & \best{\textbf{2.99$\%$}}       & \second{\textbf{4.735}}  \\
\hline
\end{tabular}
}

\vspace{-0.2in}
\end{table*}

\subsection{Shared Positional Embedding Anchor}~\label{sec: Rope}

After obtaining a high-quality subject feature {from} the decoupled image feature extractor {(Sec.~\ref{sec: Image Encoder})}, we use a \tyz{subject-aware attention} layer to fuse the subject feature with the latent feature{, which is then integrated with self-attention layer by Self-Consistent Adapter (Sec.~\ref{sec: Adapter})}. {However, we observe that when the subject feature is directly fed into the \tyz{subject-aware attention} without incorporating positional embeddings, the attention tends to disperse into regions that resemble the texture and content of the subject {(Fig.~\ref{fig: Rope Attn})}.} {This leads to a situation in the generated images where,} {although the texture and partial semantics of the subject are preserved,} {the shape and outline of the subject undergo notable changes.}

{In fact, {when calculating the cross-attention between the latent and subject features, the region outside the subject} {should be less focused,} {\textit{i.e.},} the attention {in positions {outside} the subject {region} should be suppressed.}} 
To solve this {issue} and ensure the precise utilization of the subject feature, we propose a \textbf{Shared Positional Embedding Anchor}, {which {{assembles} positional encoding in the calculation of \textit{key} to} {effectively leverage the positional information, suppressing the phenomenon of attention dispersion,} thereby {concentrating} attention in the subject area.} {Recall that the \tyz{subject-aware attention} is calculated by \szm{$\text{Cross-Attention}(Q, K_{sub}, V_{sub})$}, where \textit{query} \szm{$Q$} is calculated from the latent, and \textit{key} \szm{$K_{sub}$}, \textit{value} \szm{$V_{sub}$ from the subject features.}} We {reuse the rotary positional embedding (RoPE)} of the {base model} {and add it} to the {subject features when mapping it to the} \textit{key} {of the \tyz{subject-aware attention}}. This operation allows the interaction between the subject features and the latent space to focus on the {same} local area of the subject's contours, {mitigating} the influence of {subject} information on the global context.
\section{Experiments}
\label{sec: Exp}

\subsection{\yr{Experimental} Settings}\label{subsec: setting}

\textbf{Datasets.} 
To enable our training pipeline, we first collected 88K images of real-world photography. Among these, 19K images are classified as high-quality, characterized by their exceptional aesthetic quality and clear foreground-background relationships. Subsequently, we utilized BiRefNet~\cite{zheng2024bilateral}, ZoeDepth~\cite{bhat2023zoedepth} and OpenCV's built-in methods to generate accurate subject segmentation masks, depth maps and Sobel edge drafts, respectively. Next, we leveraged Hunyuan-Vision to create rich image descriptions, including spatial relationships and detailed descriptions of lighting, style, and background composition. Finally, we manually removed data with unreasonable descriptions or inaccurate segmentation masks. 
\\
For the evaluation, we randomly collected images of 300 everyday items. Each image is paired with the aforementioned features and three scene descriptions generated by GPT4o. Additionally, we used two random seeds for each scene description, resulting in a total of 1,800 generation tasks (300 images $\times$ 3 prompts $\times$ 2 seeds) for each method.
\\
\textbf{Implementation Details.}
\yr{We apply Pinco on two DiT backbones, Hunyuan-DiT~\cite{li2024hunyuan} and FLUX.1~\cite{flux2024}, and denote them as \textbf{HY-Pinco} and \textbf{Flux-Pinco} respectively.
For Hunyuan-DiT, we use Eq.~\ref{eq:saa} for Self-Consistent Adapter; while for FLUX.1, we use Eq.~\ref{eq: mm-attention}.
For both HY-Pinco and Flux-Pinco, we use the same shared positional embedding anchor.}
The learning rate was maintained at 0.0001, using a 1K-iteration warmup. To amplify training data, we dynamically cropped each image into different sizes, including 1:1, 16:9, and 9:16 aspect ratios. 
During evaluation, we generate images at a resolution of 1024$\times$1024 for assessment. 
We used the default parameters of each model for comparison. 
Following BrushNet~\cite{ju2024brushnet}, to ensure the visual consistency of foreground subjects, we introduced the two-stage compositing process after image generation\footnote{\footnotesize{Nearly all foreground-conditioned inpainting methods (including those in our comparison) paste the original subject region of the original image after generation.}}. Specifically, the original subject image is precisely aligned and overlaid onto the generated image within the unmasked region, preserving critical visual details while maintaining contextual coherence.
\\
\textbf{Evaluation Metrics.}
We evaluate Pinco using $8$ metrics, considering foreground consistency, text alignment, composition rationality, and image quality.
\\
1) Foreground Consistency: First, we individually measure the foreground subject's structural and perceptual consistency. To quantify structural preservation, we employ two off-the-shelf models (BiRefNet and SAM2.1-Large~\cite{ravi2024sam2}) and adopt Object-Extend-Ratio (OER)~\cite{chen2024virtualmodel} to evaluate the 
consistency between the subject shape in the generated image and the ground truth subject shape.
OER is computed as:
$\operatorname{OER} = \sum \operatorname{ReLU}(M_{s} - M_{g})/ \sum M_{g},$
where $M_{s}$, $M_{g}$ are the generated and ground truth mask of the foreground region. Since we use a two-stage compositing pipeline, a low OER metric is particularly crucial for good inpainting quality. 
Second, to assess perceptual consistency within foreground regions, we compute the LPIPS metric within the foreground subject area.
\\
2) Text Alignment: We follow Imagen3~\cite{imagenteamgoogle2024imagen3} to use VQAScore~\cite{lin2025evaluating} for alignment, given its high correlation with human assessments. In addition, we utilize the Florence-2~\cite{xiao2024florence} to measure the subject redundancy, as some models fail to establish semantic associations with the foreground region and have multiple repeated subjects.
\\
3) Composition Rationality: In many cases, it is challenging to generate images with reasonable compositions, resulting in object misalignment, imbalanced proportions, and spatial disarray. To address this, we leverage GPT-4o to evaluate each image based on three aspects: placement, size, and spatial relationships. Each image is scored on a scale of 1, 3, or 5, and the average score determines the rationality. The detailed prompt is \yr{provided in the \#Suppl}.
\\
\yr{4) Image Quality: We evaluate FID~\cite{heusel2017gans} on MSCOCO~\cite{lin2014microsoft}.}
\\
We also evaluate 
Trainable Parameters Ratio (TPR).

\begin{figure}[t]
    \centering
    \includegraphics[width=0.38\textwidth]{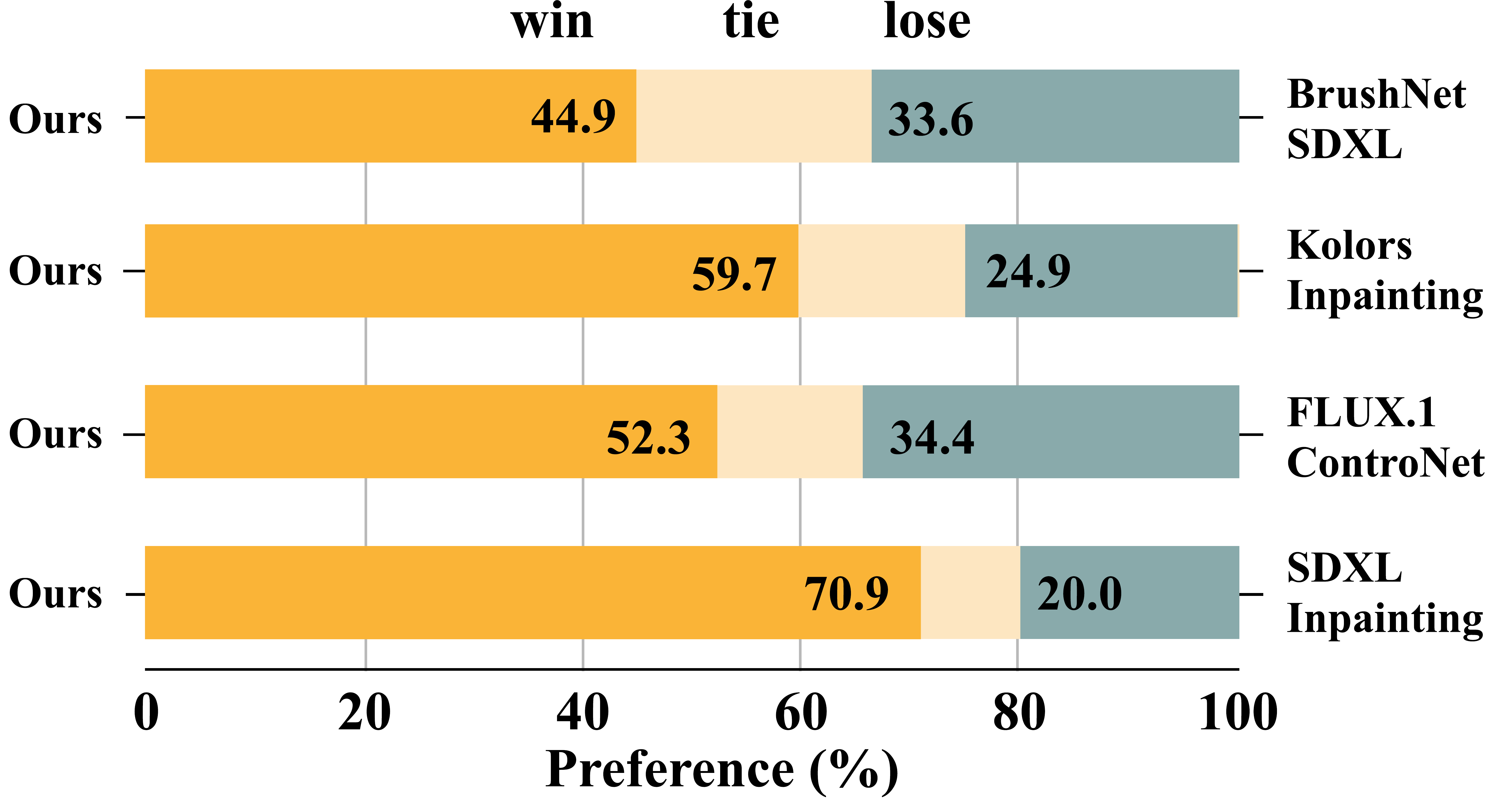}
    \caption{User Study. The images generated by Pinco received more preference from the participants compared to BurshNet~\cite{ju2024brushnet}, Kolors-IP~\cite{kolors}, Flux-IP~\cite{alimama_FLUX} and SDXL-IP~\cite{sdxl}.}
    \label{fig: User Study}
    \vspace{-0.2in}
\end{figure}

\subsection{Comparisons}

We conduct both quantitative and qualitative comparisons, along with a user study, to demonstrate the superiority of our Pinco. We quantitatively compare our Hunyuan-Pinco (HY-Pinco) and Flux-Pinco with eight UNet-based models: SD1.5 backbone (ControlNet inpainting~\cite{controlnet}, HD-Painter~\cite{manukyan2023hd}, PowerPaint~\cite{zhuang2023powerpaint}, and BrushNet-SD1.5~\cite{ju2024brushnet}), SDXL backbone (SDXL inpainting~\cite{sdxl}, layerdiffusion~\cite{li2023layerdiffusion},   BrushNet-SDXL~\cite{ju2024brushnet}, {and} Kolors-inpainting~\cite{kolors})   
{as well as} two DiT-based models: HY-Controlnet, and Flux ControlNet~\cite{alimama_FLUX}. 
Additionally, we add Adobe Firefly~\cite{firefly2023} in qualitative comparison.

\noindent
\textbf{Quantitative Comparison.}
As shown in Tab.~\ref{tab: Metric Comp}, even with a smaller number of learning parameters, Pinco achieved leading results across various evaluation metrics. Our HY-Pinco and Flux-Pinco outperform others, notably in OER scores, demonstrating exceptional ability to maintain the subject's shape integrity and preserve contours and details. On the same base model, compared to ControlNet, Pinco shows a higher VQA-Score and a lower FV2Score. This highlights its outstanding capabilities in following the contents of the input prompts. Finally, GPT4o awarded Pinco the highest rationality score, showing its effectiveness in accurately managing the spatial relationship between the given foreground subject and the generated content.

\noindent
\textbf{Qualitative Results.} The qualitative comparison with previous image inpainting methods is illustrated in Fig.~\ref{fig: vision comp}. In foreground-conditioned inpainting tasks that involve complex textual descriptions, many methods neglect to draw certain objects in the background 
(such as the beach chairs referenced in the prompt for the images in the 5th row). Additionally, some approaches face challenges with subject expansion, a problem particularly noticeable with the vacuum cleaner in the 6th row. Most methods also fail to adequately consider the spatial relationships between the subject and the background, leading to subjects appearing disproportionately large or mispositioned within the scene (as seen with the oversized humidifier on the bed in the 4th row). In contrast, our HY-Pinco and Flux-Pinco effectively generate coherent and detailed backgrounds that align with the provided text while successfully mitigating the issue of subject expansion, achieving enhanced image synthesis quality. These results substantiate the strong generalization capability of the Pinco adapter when implemented with different DiT-based architectures.
More comparisons can be found in the \#Suppl.


\begin{table}[t]
\small
\caption{Ablation quantitative comparison on Pinco modules.}
\vspace{-0.05in}
\centering
\setlength{\abovecaptionskip}{-0.4cm}
\renewcommand{\arraystretch}{1.1}
\resizebox{1.0\linewidth}{!}{
\begin{tabular}{c|ccccccc}
\hline
Models         & TPR $\downarrow$    & FID $\downarrow$   & VQAScore $\uparrow$ & OER(SAM2.1) $\downarrow$ & OER(BiRefNet) $\downarrow$ & FV2Score $\downarrow$ & GPT4o $\uparrow$\\ \hline
Pinco-wo-RoPE & 10.46\%  & 86.13 & 0.892    & 290.44\%     & 68.64\%       & 7.37\%   & \textbf{4.820} \\
Pinco-vae-only & \textbf{9.48\%}  & 86.13 & 0.889    & 36.11\%     & 14.58\%       & 8.15\%   & 4.688 \\
Pinco-Cross    & 11.37\% & 85.61 & 0.892    & 14.55\%     & 10.24\%       & 3.51\%   & 4.745 \\
Pinco-Self (ours)     & 11.37\% & \textbf{84.25} & \textbf{0.901}    & \textbf{11.51\%}     & \textbf{10.00\%}       & \textbf{3.16\%}   & 4.790  \\ \hline
\end{tabular}
}
\label{tab: self}
\vspace{-0.1in}
\end{table}

\begin{figure}[t]
    \centering
    \includegraphics[width=0.4\textwidth]{sec/assets/Ablation_Comp.pdf}
    \caption{Qualitative Ablation Comparisons of Pinco, Pinco-Cross, Pinco-VAE and Pinco-w/oRoPE.}
    \label{fig: vision ab}
    \vspace{-0.2in}
\end{figure}

\noindent
\textbf{User Study.} 
We compared with four representative competitors for this user study. 31 participants took part in our user study, with each participant evaluating 40 sets of questions. Each set included two {randomly arranged side-by-side images} generated by Pinco and another method. Participants were asked to select the superior image based on several criteria, including {foreground consistency} preservation, text alignment, and overall image quality. The user voting results in Fig.~\ref{fig: User Study} demonstrate that Pinco is more favored by participants, as our model consistently received higher preference scores across all comparisons, indicating its effectiveness in producing high-quality results.

\subsection{Ablation Study}~\label{sec: ablation study}
To validate the effectiveness of the proposed key components, we perform extensive ablation studies on HY-Pinco.

\noindent
\textbf{Impact of Self-Consistent Adapter Injection.}
Fig.~\ref{fig: vision ab} shows that injecting subject features into the cross-attention layer can lead to three significant issues: {text-driven} shape expansion (illustrated in the first row of images), illogical spatial arrangements (notably in the second row, where the washing liquid appears to float), and inadequate text alignment (as seen in the third row, {which fails to include the ``scattered glasses" mentioned in the prompt}). The quantitative analysis presented in Tab.~\ref{tab: self} further highlights the advantages of incorporating in the self-attention layer, effectively mitigating shape expansion and ensuring a more coherent output. {This reinforces the idea that self-attention is more aligned with the overall image layout.} {Furthermore, our empirical observations indicate that self-attention injection significantly boosts training efficiency and leads to much faster convergence compared to the cross-attention layer.} More results are presented in the \#Suppl.

\begin{figure}[t]
    \centering
    \includegraphics[width=0.45\textwidth]{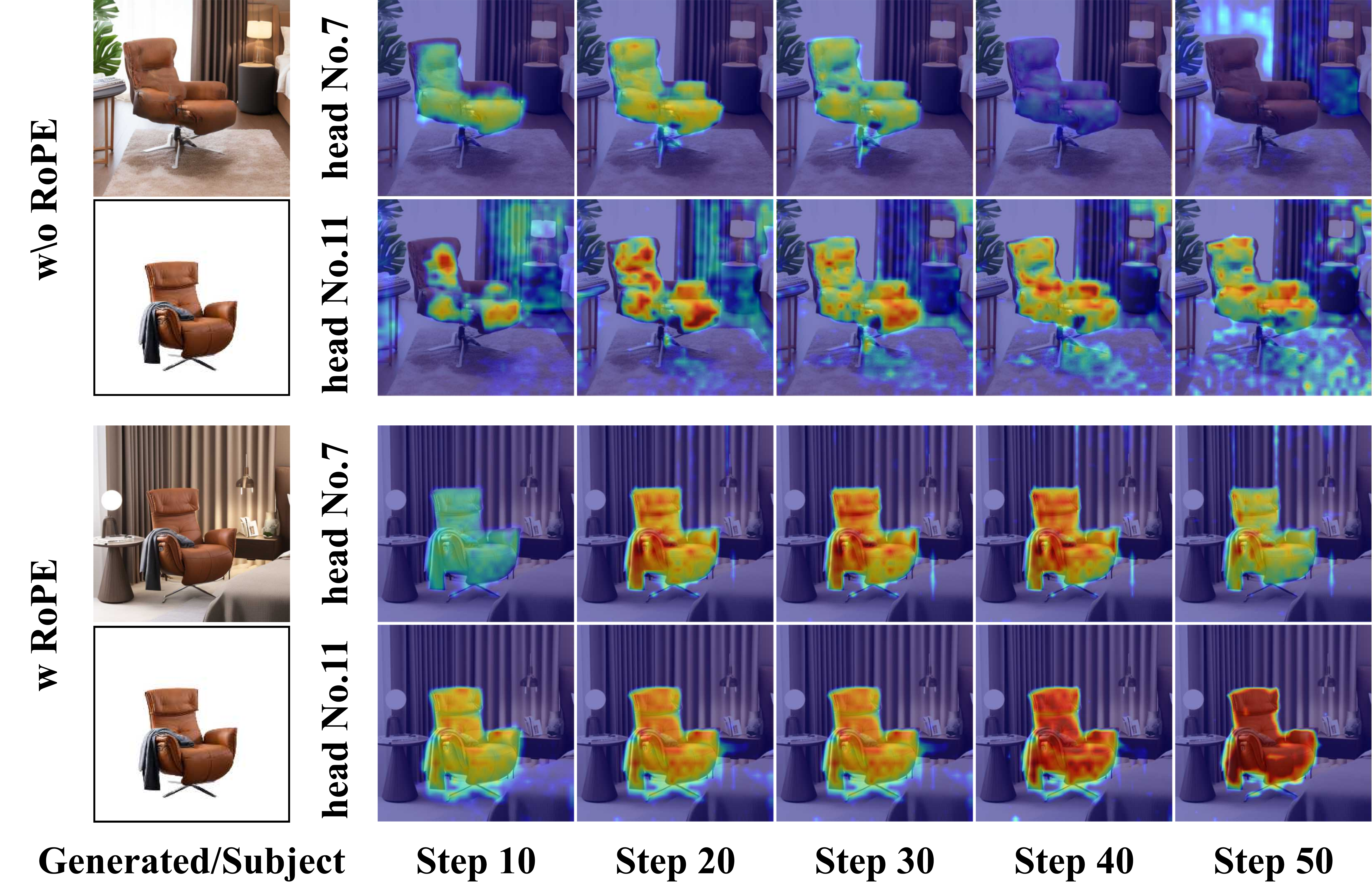}
    \caption{The attention maps of the model with rope and the model without rope. As the inference progresses, the model with RoPE gradually focuses its attention on the main subject, while the model without RoPE tends to focus on an incorrectly shaped subject and its attention map is scattered.}
    \label{fig: Rope Attn}
    \vspace{-0.2in}
\end{figure}

\noindent
\textbf{Impact of Decoupled Image Feature Extractor (DIFE).}
The comparison of the expansion rate metric is presented in Tab.~\ref{tab: self}, {highlighting the performance difference between the Variational Autoencoder (VAE) and the Decoupled Image Feature Extractor (DIFE) as feature encoder.} Notably, the use of DIFE results in a significant reduction in the {OER}(SAM2.1), decreasing it by 24.6\% compared to the VAE approach. Also, the OER(BiRefNet) is reduced by 4.58\% when employing DIFE. Furthermore, the visual comparison in Fig.~\ref{fig: vision ab} reveals that relying solely on VAE for extracting subject features tends to have noticeable shape expansion. This observation confirms the effectiveness of DIFE in mitigating such issues, highlighting its advantages in maintaining the integrity of the extracted features.

\begin{figure}[t]
    \centering
    \includegraphics[width=0.9\linewidth]{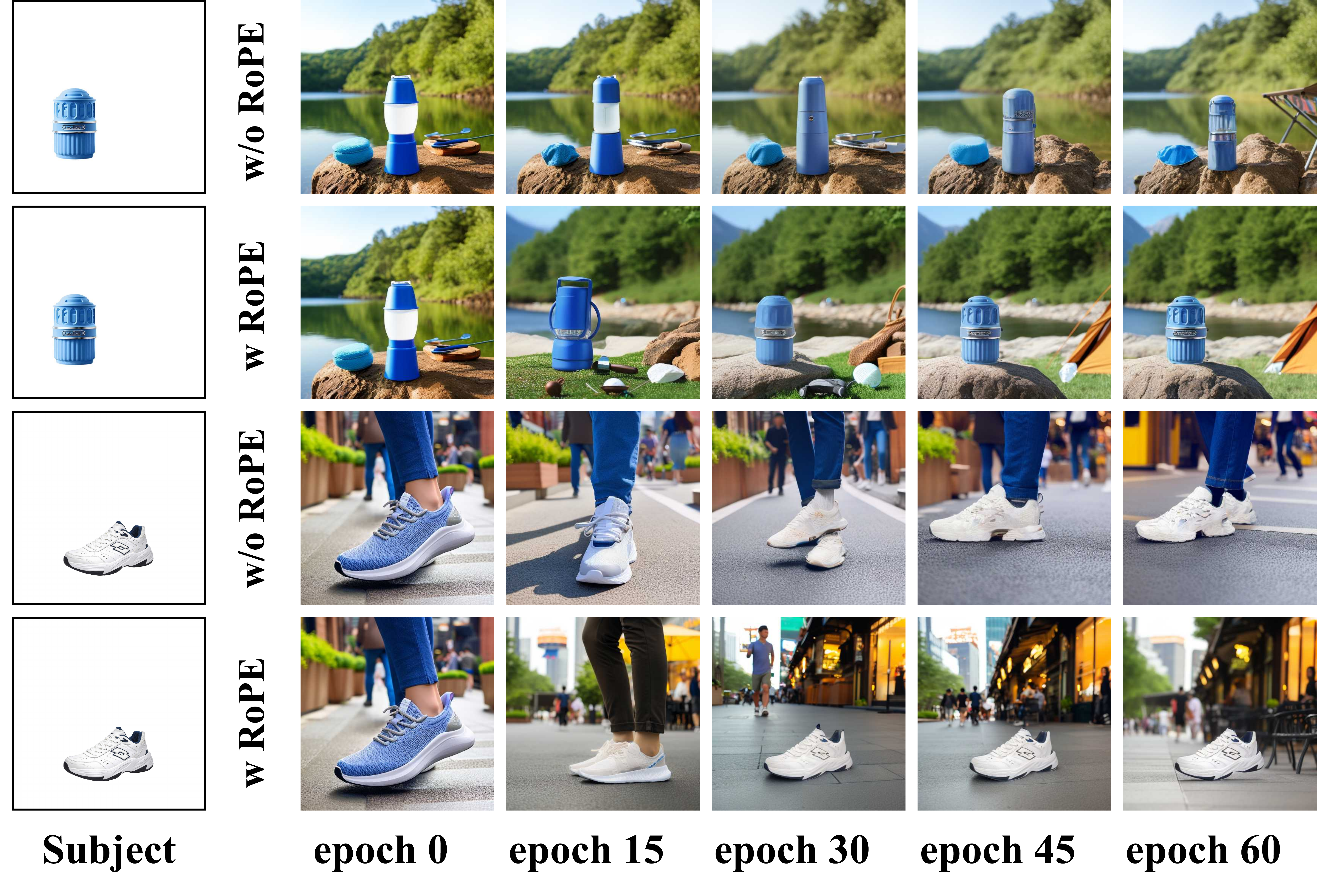}
    \vspace{-0.05in}
    \caption{The convergence process of Pinco with RoPE and without RoPE. Without the help of a positional embedding anchor, the model fails to converge and can not generate an image of the correct subject.}
    \label{fig: Rope Comp}
    \vspace{-0.2in}
\end{figure}

\noindent
\textbf{Impact of Shared Positional Embedding Anchor.}
Fig.~\ref{fig: vision ab} illustrates that in the absence of a shared positional embedding anchor, the generated images primarily retain the subject's color distribution and abstract semantic information, while the subject's shape and texture are largely compromised. A closer examination of the attention maps in Fig.~\ref{fig: Rope Attn} reveals a phenomenon of attention dispersion when position embedding is not utilized. This suggests that the model tends to focus on global features, neglecting the local shape details of the subject. In contrast, the attention map of the model that incorporates the shared positional embedding anchor is well-aligned with the subject's outline. Additionally, as shown in Fig.~\ref{fig: Rope Comp}, the model with the positional embedding anchor effectively concentrates on the subject area when injecting subject features, leading to a significant improvement in consistency and convergence efficiency.

\section{Conclusion}\label{sec: Col}
In this paper, we introduce Pinco, a novel plug-and-play adapter for {diffusion transformers in}  foreground-conditioned inpainting. Pinco effectively {addresses} the challenges of generating high-quality backgrounds while preserving the {consistency} of the foreground subject. {Our proposed} Self-Consistent Adapter facilitates a harmonious interaction between foreground features and the overall image layout, mitigating conflicts that can arise from text and subject discrepancies. The Decoupled Image Feature Extraction method {leverages distinct architectures to capture} both semantic and shape features, {achieving improved feature extraction} and fidelity in shape preservation. Additionally, the Shared Positional Embedding Anchor allows for a more precise focus on the subject region, further enhancing the model's performance and training efficiency. Extensive experimental results confirm that Pinco outperforms existing methods, providing a robust solution for high-quality foreground-conditioned inpainting with good text alignment and subject shape preservation.

\section*{Acknowledgments}
This work was supported by National Natural Science Foundation of China (No. 62302297, 72192821, 62272447, 62472282, 62472285), Young Elite Scientists Sponsorship Program by CAST (2022QNRC001), the Fundamental Research Funds for the Central Universities (YG2023QNB17, YG2024QNA44), Beijing Natural Science Foundation (L222117), Tencent Marketing Solution Rhino-Bird Focused Research Program (Tencent AMS RBFR2024005).
{
    \small
    \bibliographystyle{ieeenat_fullname}
    \bibliography{main}
}
\clearpage
\appendix
\section*{\LARGE Appendix}

\section{Overview}
In this supplementary material, we mainly present the following components:
\begin{itemize}
    \item More implementation details of our model structure and training, and more qualitative results in Sec.~\ref{sec: id}.
    \item More ablation study on the convergence of training in Sec.~\ref{sec: AS}.
    \item More details and cases of the user study in Sec.~\ref{sec: US}.
    \item More qualitative comparisons between our Pinco and the state-of-the-art methods in Sec.~\ref{sec: QC}.
    \item More details and results of the GPT-4o rationality analysis in Sec.~\ref{sec: gpt}.
    \item More inference results under special cases in Sec.~\ref{sec: SC}.
    \item Limitations in Sec.~\ref{sec: limit}
    \item Image copyright in Sec.~\ref{sec: cpr}
\end{itemize}

\section{Implementation Details}\label{sec: id}
\subsection{Model Architecture}

\begin{figure*}[t]
    \centering
    \includegraphics[width=0.98\textwidth]{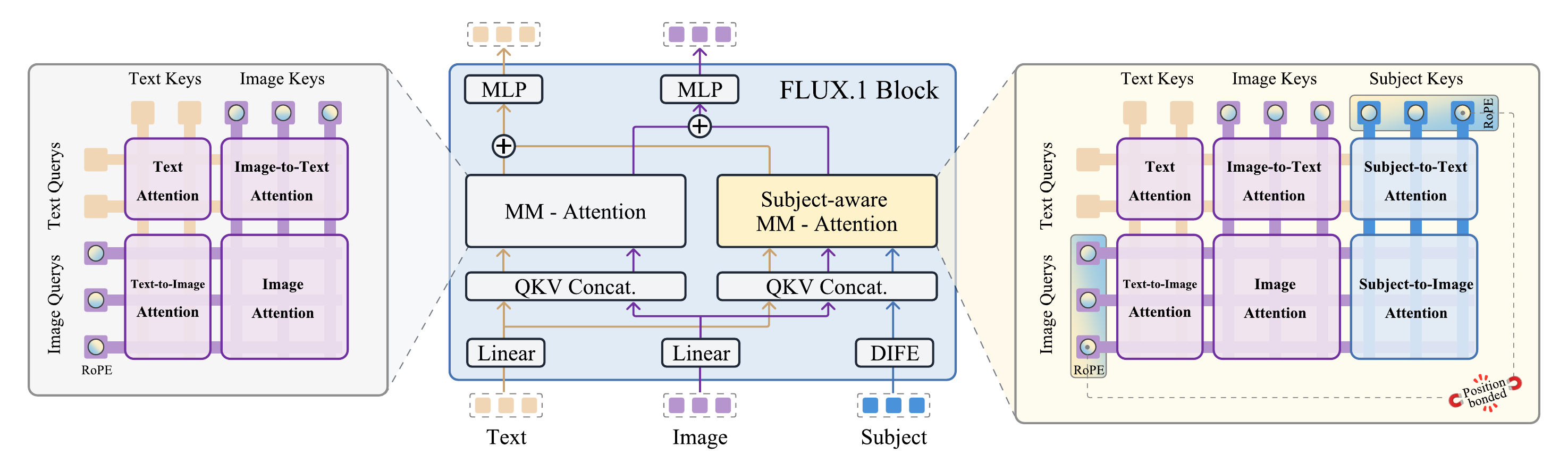}
    \caption{The architecture of Flux-Pinco. For MM-DiT, we concatenate the latent and the subject features together to calculate the subject-aware attention.}
    \label{fig: Flux-Pinco}
\end{figure*}

\textbf{Backbone.} We apply our proposed Pinco on two DiT-based models, Hunyuan-DiT and Flux-DiT models. For Hunyuan-DiT~\cite{li2024hunyuan}, we use the $DiT-g/2$ config which consists of $40$ blocks and has a $1,408$ embed\_dim.
For Flux-DiT, our Pinco is applied to both the DoubleStreamBlocks and the SingleStreamBlocks. The architecture of Flux-Pinco is shown in Fig.~\ref{fig: Flux-Pinco}.

\noindent
\textbf{Decoupled Image Feature Extractor.} We use the VAE Encoder of the original DiT model as our semantic feature extractor and only take the output of the final layer as our semantic feature. Meanwhile, we also construct a simple convolutional network (ConvNet) to extract the shape feature. 
More precisely, since Hunyuan-DiT draws inspiration from the ideas of U-ViT~\cite{bao2023worthwordsvitbackbone} by using skip connections to link the blocks of DiT, we believe that the presence of skip connections allows different blocks to have varying granularities. If we directly feed the output of the last layer of the ConvNet into all modules without distinction, it would weaken the model's inherent perception of feature granularity. Therefore, we extract the outputs from different layers of the ConvNet for different blocks. Specifically: 
\begin{itemize}
    \item The ConvNet consists of 7 convolutional layers, with the outputs of the 1st, 3rd, 5th, and 7th convolutional layers serving as features.
    \item For Hunyuan-DiT, we divided the 40 blocks into 8 groups, with each group containing 5 blocks. Considering the skip connections, we feed the features from the first layer of the ConvNet into blocks 1-5 and 36-40, and the features from the second layer into blocks 6-10 and 31-35, and so on.
    \item For Flux-DiT, we apply our pinco on both the 19 DoubleStreamBlocks and the 38 SingleStreamBlocks. And the use of the ConvNet is the same to the HY-Pinco.
\end{itemize}

\begin{figure}[t]
    \centering
    \includegraphics[width=0.47\textwidth]{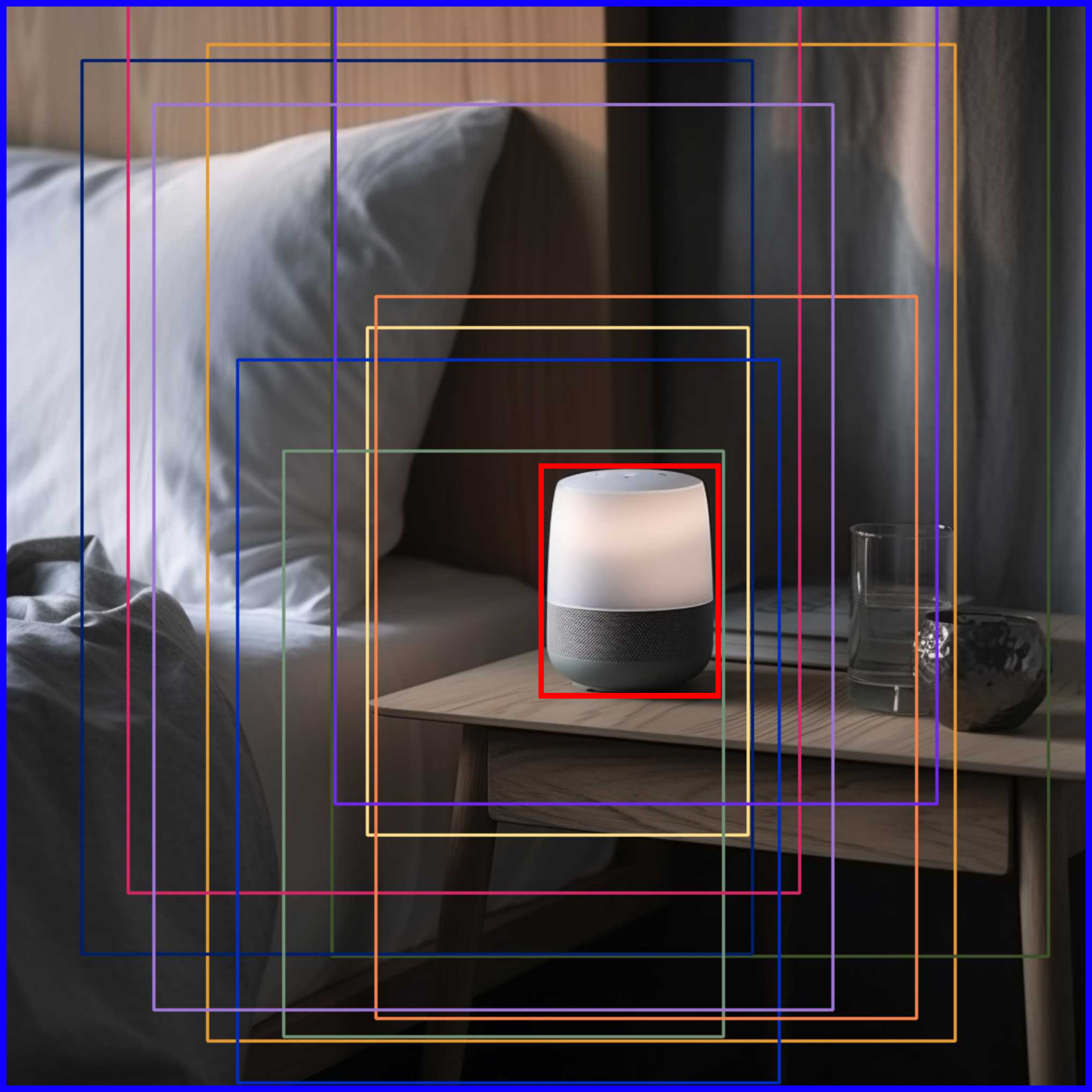}
    \caption{Method for obtaining multi-aspect ratio samples.}
    \label{fig: multi fram}
\end{figure}

\noindent
\textbf{Self-Consistent Adapter.}  We construct corresponding adapters for each block to inject the subject feature. Each adapter first rearranges the feature shape to match the intrinsic latent shape~\cite{xue2024stldmuniversalframeworktextgrounded} and then uses a linear layer to transform features from a dimension of $dim$ to a dimension of the latent feature. Then, it uses two independent matrices to obtain $K$ and $V$, which are used to compute the subject-aware attention. The $Q$ is directly taken from the model's computation of self-attention. 

\begin{figure*}[!t]
    \centering
    \includegraphics[width=0.8\textwidth]{sec/assets/GPT4O_QA.pdf}
    \caption{GPT-4o prompt for assessment and its reply. Note that you need to specify the name of the subject in \second{[subject]}}
    \label{fig: gpt_4o_qa}
\end{figure*}

\subsection{Multi-Aspect Ratio Training}

Due to the varying proportions of subjects within the frame (e.g., a vehicle that may occupy more than 50\% of the frame, while an item like a shoe might only take up about 10\%), it is essential for the model to effectively handle 
subjects occupying different proportions of the frame
and generate a suitable background for subjects of varying sizes. 
To achieve this, we employ a multi-aspect ratio augmentation method to construct training samples throughout the training process. As illustrated in Fig.~\ref{fig: multi fram}, for each high-quality image, we define the minimum bounding rectangle of the subject based on the mask (in \best{red}), while the bounding rectangle of the entire image serves as the maximum range (in \second{blue}). We designate the areas of the maximum and minimum frames as the upper and lower bounds of a normal distribution, respectively. During training, we sample various shapes and locations of bounding rectangles to create training samples with diverse frame proportions and aspect ratios (e.g., 16:9, 9:16, 1:1). 
Fig.~\ref{fig: vs} shows more cases generated by Pinco with different aspect ratios.

\section{More Ablation Study}\label{sec: AS}
\subsection{Convergence Process Analysis}
\begin{figure}[t]
    \centering
    \includegraphics[width=0.47\textwidth]{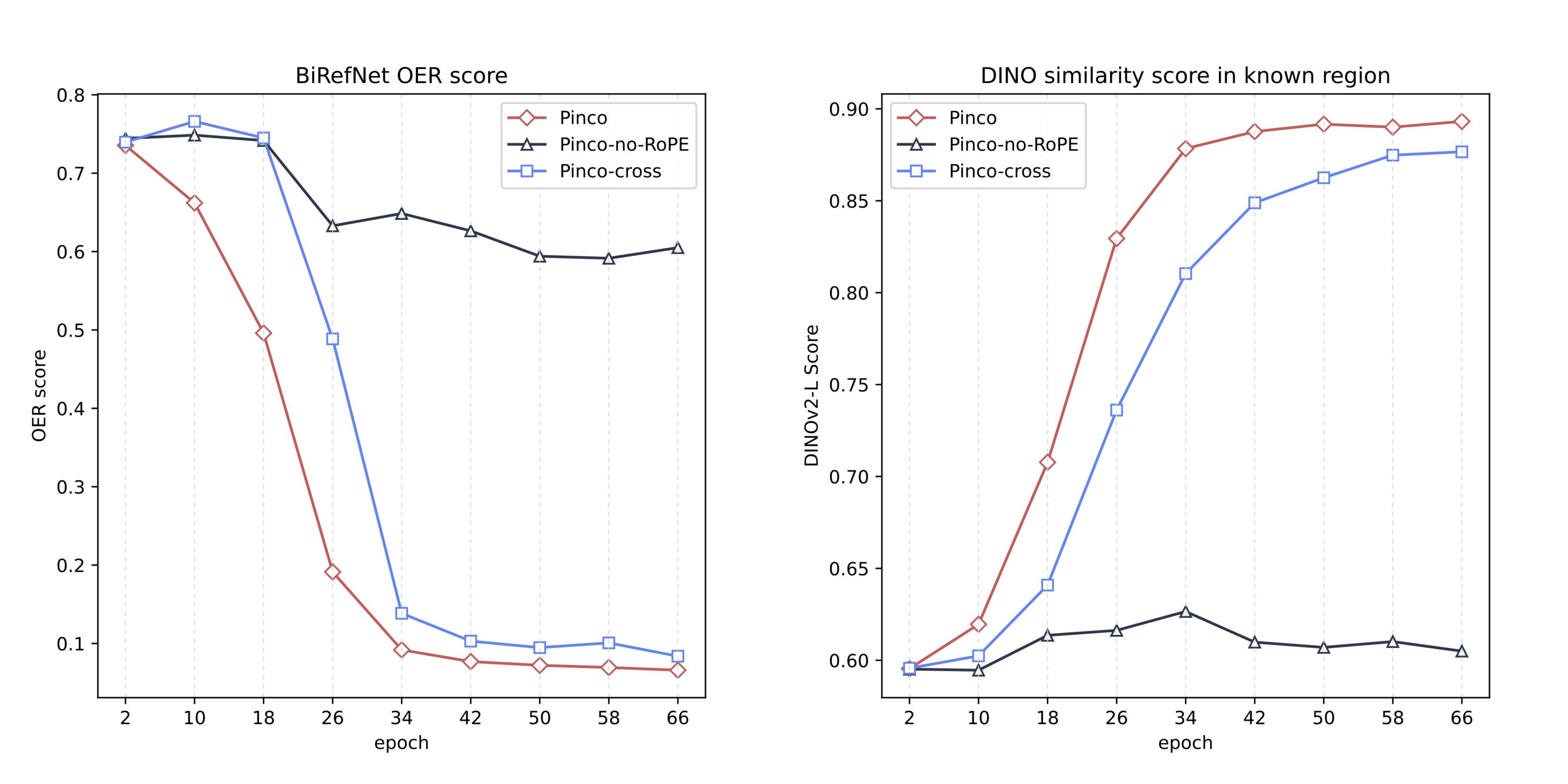}
    \caption{The convergence analysis of Pinco, Pinco-Cross, and Pinco-w/oRoPE. Pinco-Self can maintain better shape constraints and foreground consistency while achieving efficient training.}
    \label{fig: self Comp}
     \vspace{-0.2in}
\end{figure}

Fig.~\ref{fig: self Comp} shows the convergence analysis of HY-Pinco, Pinco-w/oRoPE, and Pinco-Cross. The two images illustrate the changes in image OER and DINO similarity within the mask area as the training epochs extended. Evidently, without the aid of shared positional embedding anchor, the model struggles to effectively incorporate subject features, resulting in consistently poor scores for both OER and DINO similarity. This highlights the importance of shared positional encoding in effectively utilizing subject features and ensuring the consistency of the subjects in the generated images. On the other hand, injecting subject features in the self-attention layer leads to faster convergence during training, with lower OER and better DINO similarity compared to injecting features in the cross-attention layer. This further supports the rationale and effectiveness of injecting features in the self-attention layer. Fig.~\ref{fig: rope} demonstrates more cases between Pinco and Pinco-w/oRoPE during the training process.

\section{User Study}\label{sec: US}

During the user study, participants were asked to evaluate side-by-side samples from multiple aspects, including the rationality of the background, the appropriateness of object sizes, the suitability of object placements, and the harmony between the subjects and the background, and select the images they considered to be better. Fig.~\ref{fig: user_s} displays some cases from the user study.

\section{More Qualitative Comparisons}\label{sec: QC}
We provide more qualitative comparisons between our Pinco and the state-of-the-art methods in Fig.~\ref{fig: qu_1},~\ref{fig: qu_2} and \ref{fig: qu_3}.
The compared baselines include:
\begin{itemize}
    \item SD1.5 backbone: ControlNet inpainting~\cite{controlnet}, HD-Painter~\cite{manukyan2023hd}, PowerPaint~\cite{zhuang2023powerpaint}, and BrushNet-SD1.5~\cite{ju2024brushnet};
    \item SDXL backbone: SDXL inpainting~\cite{sdxl}, layerdiffusion~\cite{li2023layerdiffusion},   BrushNet-SDXL~\cite{ju2024brushnet}, {and} Kolors-inpainting~\cite{kolors};
    \item DiT-based models: HY-ControlNet (Hunyuan-DiT backbone), and Flux ControlNet~\cite{alimama_FLUX} (FLUX.1 backbone).
\end{itemize}

\section{GPT-4o Rationality}\label{sec: gpt}
We leverage the GPT-4o to evaluate each image based on Object Placement Relationship, Object Size Relationship, and Physical Space Relationship. The criteria for these three aspects and rating criteria are as follows:
\begin{itemize}
    \item Object Placement Relationship: Check whether the spatial relationship between the subject and other objects in the image is reasonable and consistent with common placement methods in daily life. Determine if the subject is placed in a physically impossible position, such as floating.
    \item Object Size Relationship: Assess whether the size proportions between the subject and other objects in the image are realistic and whether there is any disproportion between the subject and surrounding objects.
    \item Physical Space Relationship: Consider whether the spatial distance between the subject and other objects in the image is reasonable, whether the perspective relationship conforms to the laws of the physical world, and whether there are any unreasonable aspects.
    \item Rating Criteria: 1 point: Obvious errors, inconsistent with the real world. 3 points: Minor errors, somewhat inconsistent with the real world. 5 points: No obvious errors, consistent with the real world.
\end{itemize}

Fig.~\ref{fig: gpt_4o} presents the results of the rationality analysis for the images returned by GPT-4o under the criteria mentioned above. The detailed prompt given to GPT-4o and the reply are shown in Fig.~\ref{fig: gpt_4o_qa}.

\section{Inference Results under Special Cases.}\label{sec: SC}
\begin{figure}[t]
    \centering
    \includegraphics[width=\linewidth]{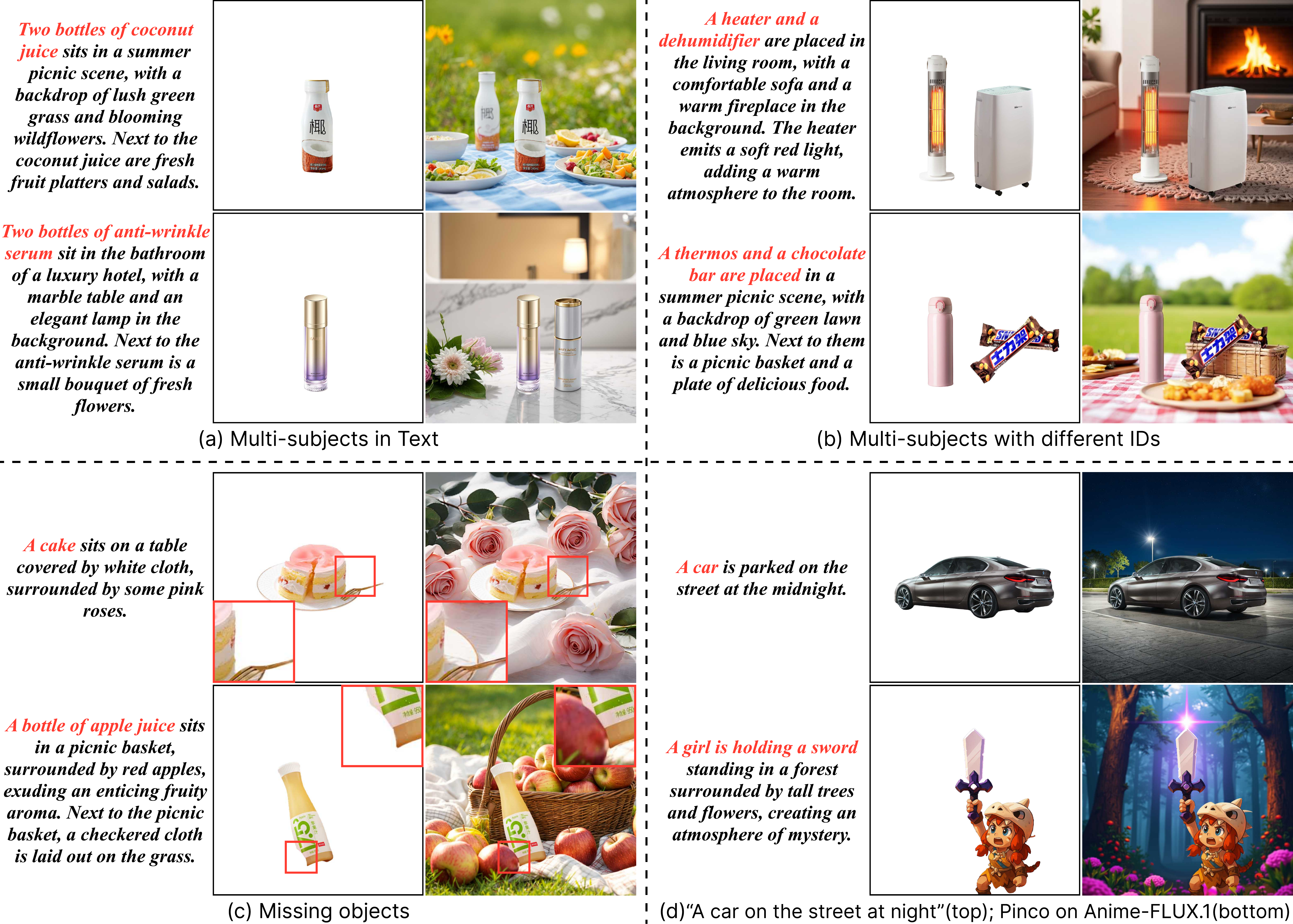}
    \caption{Some inference results under special cases.}
    \label{fig: rebuttal}
\end{figure}

To verify the robustness of our method in the inference phase, we conducted the following special experiments:

\noindent \textbf{Text-image Interdependence.} We test the case where the number of subjects in the text description is greater than the number of subjects in the conditional image.
As shown in Fig.~\ref{fig: rebuttal} (a), our Pinco will generate another subject which is aligned with the given prompts.
It is worth noting that additional subjects appear only when the text describes multiple subjects.

\noindent \textbf{Multiple Subjects with Different IDs.} We test the case where there are multiple subjects with different IDs in the conditional foreground image.
As shown in Fig.~\ref{fig: rebuttal} (b), our Pinco can generate perfect results aligned with the given prompts.

\noindent \textbf{Completing Missing Objects. } We test the case where the conditional foreground objects have missing parts. 
As shown in Fig.~\ref{fig: rebuttal} (c), for partially missing objects, our Pinco either completes the missing foreground parts or uses background objects to cover them for a harmonious result.

\noindent \textbf{Textual Conflicts.} We test the case where the conditional foreground image and the textual background description conflict in lighting conditions.
In such case, the model will adjust the background to better fit the foreground while ensuring alignment with the text. For example, given a foreground car in bright scene and the textual description of night, the result might show a car lit by a street lamp on a nighttime street, As shown in Fig.~\ref{fig: rebuttal} (d) (top).

\noindent \textbf{Plug and Play Property.} As an adapter, our Pinco can be applied to different models with the same structure with no need of extra training. For example, we apply our Pinco trained on FLUX.1-dev onto an Anime-style-finetuned FLUX by community. As shown in Fig.~\ref{fig: rebuttal} (d) (bottom), it can still generate excellent foreground-conditioned inpainting results.

\section{Limitations}\label{sec: limit}
In foreground-conditioned background inpainting task, one of the most significant requirements is to ensure the foreground consistency.
Existing methods generally maintain the internal detailed features of the input foreground well.
However, due to their unstable feature injection, the model usually generates some extended parts based on its hallucinations.
In Pinco, although we proposed dedicated Self-Consistent Adapter to facilitate a harmonious interaction between foreground features and the overall image layout, sometimes it is still hard to control the shape when dealing with very slender objects like ropes or sticks.
In addition, when the input foreground object is captured from an unusual viewpoint, the model may not understand the perspective of the object, resulting in an unreasonable positional relationship between the generated background and foreground.

\section{Copyright}\label{sec: cpr}
Some of the images presented in this paper are sourced from publicly available online resources. In our usage context, we have uniformly retained only the main subject of the original images and removed the background parts. In this section, we specify the exact sources of the images in the form of image links and author credits in Tab.~\ref{tab: copyright}. Except for the images explicitly credited with the author and source, all other images are derived from client cases or generated by open-sourced models. The copyright of the images belongs to the original authors and brands. \textbf{The images used in this paper are solely for academic research purposes and are only used to test the effectiveness of algorithms. They are not intended for any commercial use or unauthorized distribution.}

For each image containing multiple sub-images, we number them in order from left to right and from top to bottom. For example, in Figure.~\ref{fig: vs}, the sub-images in the first row are numbered as SubFig.S6-1, SubFig.S6-2, and so on, while the sub-images in the second row are numbered as SubFig.S6-5, and so forth.

\begin{table}[h]
\scriptsize
\centering
\setlength\tabcolsep{2pt}
\renewcommand{\arraystretch}{1.1}
\caption{Copyright of the images in our paper.}
\label{tab: copyright}
\resizebox{\columnwidth}{!}{%
\begin{tabular}{l|l}
\toprule
Figure & Source \\
\midrule
\multirow{3}*{\textbf{Figure.~1}} & \href{https://scontent-hkg4-1.xx.fbcdn.net/v/t39.30808-6/424524678_372133392473459_8853722487271008359_n.jpg?_nc_cat=106\&ccb=1-7\&_nc_sid=f727a1\&_nc_ohc=CaoMpHba4cIQ7kNvgFGCanI\&_nc_zt=23\&_nc_ht=scontent-hkg4-1.xx\&_nc_gid=AN1f0kMjvYy10FgZlyKfKAX\&oh=00_AYC9q700UgP1h5gUiKeHCQDbziL6G2F4sd1eoXIVCPu8rQ\&oe=674E11E0}{SubFig.1-2} (ANTA), \href{https://www.zcool.com.cn/work/ZNTkyNTMzMDg=.html}{SubFig.1-3} (METHOU), \\
& \href{https://us.balmuda.com/}{SubFig.1-5} (BALMUDA), \href{http://www.taolinzhen.com/upload/tmp/TB29NFqfGmWBuNjy1XaXXXCbXXa_!!763968012.jpg}{SubFig.1-6} (FLYCO), \\
&\href{https://images8.alphacoders.com/919/thumb-1920-919688.jpg}{SubFig.1-7} (Alpha Coders)\\
&\\
\textbf{Figure.~2} & \href{https://www.zcool.com.cn/work/ZNjY2MjU1NzY=.html}{SubFig.2-1} (ZCOOL) \\
&\\
\textbf{Figure.~3} & SubFig.3-1 (Mazda) \\
&\\
\multirow{3}*{\textbf{Figure.~4}}  & \href{https://www.zcool.com.cn/work/ZNTA4NTc0MDg=.html}{SubFig.4-1} (Foodography), \href{https://www.zcool.com.cn/work/ZNTI4NDEzODQ=.html}{SubFig.4-2} (Xiangma), \\
& \href{https://www.zcool.com.cn/work/ZNTY4MDAwOTI=.html}{SubFig.4-3} (IMAXER), \href{https://www.zcool.com.cn/work/ZMzgzMTgzODQ=.html}{SubFig.4-4} (Foodography), \\ 
& {SubFig.4-5} (IM Motors), \href{https://www.zcool.com.cn/work/ZMzk3NjA2ODA=.html}{SubFig.4-6} (Foodography) \\
&\\
\textbf{Figure.~6} & \href{https://www.zcool.com.cn/work/ZNTU5Mjk4OTY=.html}{SubFig.6-1} (NOW VISION), \href{https://www.zcool.com.cn/work/ZNjA5OTk0MzY=.html}{SubFig.6-2} (Foodography) \\
&\\
\textbf{Figure.~7} & \href{https://www.zcool.com.cn/work/ZMzUzNDA2NDQ=.html}{SubFig.7-2/4} (MANTO) \\
&\\
\textbf{Figure.~8}  & \href{https://gd-hbimg.huaban.com/8b4506202fd5aecd15c6ccb0485ccd3a969ccdc58b20e-e6X6ax_fw658}{SubFig.8-1/2} (Foodography), \href{https://gw.alicdn.com/imgextra/i4/2218297282346/O1CN01JKuMZB1TCW9KFpeUH_!!2-item_pic.png_468x468Q75.jpg_.webp}{SubFig.8-3/4} (LOTTO) \\
&\\
\textbf{Figure.~\ref{fig: gpt_4o_qa}} & \href{https://img.zcool.cn/community/66a664ed007dbqg3a18me65706.jpg?imageMogr2/format/webp}{SubFig.S3-1/2} (DESING) \\
&\\
\multirow{5}*{\textbf{Figure.~\ref{fig: rebuttal}}}  & \href{https://www.zcool.com.cn/work/ZNjE2ODE0MTY=.html}{SubFig.S5-a1} (Foodography), \href{https://www.zcool.com.cn/work/ZNTE3Mzk5Njg=.html}{SubFig.S5-a2} (Foodography), \\
& \href{https://www.zcool.com.cn/work/ZNDgzMzM5MTI=.html}{SubFig.S5-b1/1} (Foodography), \href{https://www.zcool.com.cn/work/ZNDUzMDUyNDQ=.html}{SubFig.S5-b1/2} (Foodography), \\ 
& \href{https://you.163.com/item/detail?id=3987747}{SubFig.S5-b2/1} (Lifease), {SubFig.S5-b2/2} (Snickers), \\ 
& SubFig.S5-c1 (Cake),  \href{https://www.zcool.com.cn/work/ZNjY0NjAzNjA=.html}{SubFig.S5-c2} (Box Studio), \\
& \href{https://c4.wallpaperflare.com/wallpaper/969/438/447/2015-bmw-compact-sedan-concept-wallpaper-preview.jpg}{SubFig.S5-d1} (Wallpaper Flare), \href{https://yblb.feiyu.com/}{SubFig.S5-d2} (Feiyu), \\
&\\
\multirow{9}*{\textbf{Figure.~\ref{fig: vs}}}  & \href{https://www.zcool.com.cn/work/ZNTg4ODc4ODQ=.html}{SubFig.S6-1} (MAOOXD), \href{https://www.zcool.com.cn/work/ZNTA4MDIzODQ=.html}{SubFig.S6-2} (Foodography), \\
& \href{https://www.zcool.com.cn/work/ZNDQ5NjQ4MTY=.html}{SubFig.S6-3} (Foodography), {SubFig.S6-4} (SUPOR), \\ 
& \href{https://www.zcool.com.cn/work/ZNzA1MTUyODg=.html}{SubFig.S6-5} (JianmuPhotography), \href{https://images4.alphacoders.com/940/thumb-1920-940370.jpg}{SubFig.S6-7} (Alpha Coders), \\ 
& SubFig.S6-8 (BYHEALTH),  \href{http://xhslink.com/a/vgZIQxwJthB0}{SubFig.S6-9} (Rarakiddo), \\
& \href{https://www.zcool.com.cn/work/ZNDEzOTQ3NTI=.html}{SubFig.S6-10} (Foodography), \href{https://us.balmuda.com/}{SubFig.S6-11} (BALMUDA), \\
& SubFig.S6-12 (LIBY), \href{http://xhslink.com/a/27ADODsOqVF0}{SubFig.S6-13} (ROLEX), \\ 
& SubFig.S6-14 (LUXEED), \href{https://imgservice.suning.cn/uimg1/b2c/image/egtrzYVVInCrSGNk7IayBQ.jpg_800w_800h_4e}{SubFig.S6-15} (L'Oreal), \\
& \href{https://unsplash.com/photos/pepsi-soda-tin-can-ElfJDs4LAQk}{SubFig.S6-16} (Olena Bohovyk), \href{https://us.balmuda.com/}{SubFig.S6-17} (BALMUDA), \\
& \href{https://us.balmuda.com/}{SubFig.S6-18} (BALMUDA) \\
&\\
\multirow{5}*{\textbf{Figure.~\ref{fig: rope}}} & \href{https://www.zcool.com.cn/work/ZNTI4NDEzODQ=.html}{SubFig.S7-1} (Xiangma), \href{https://www.zcool.com.cn/work/ZNTkyNTMzMDg=.html}{SubFig.S7-2} (METHOU), \\ 
& \href{https://www.zcool.com.cn/work/ZNTE3Mzk5Njg=.html}{SubFig.S7-3} (Foodography), \href{https://www.zcool.com.cn/work/ZNTU5Mjk4OTY=.html}{SubFig.S7-4} (NOW VISION), \\
& SubFig.S7-5 (Mazda), \href{https://www.zcool.com.cn/work/ZNzA1MTUyODg=.html}{SubFig.S7-6} (JianmuPhotography), \\ 
& \href{https://scontent-hkg4-1.xx.fbcdn.net/v/t39.30808-6/424524678_372133392473459_8853722487271008359_n.jpg?_nc_cat=106\&ccb=1-7\&_nc_sid=f727a1\&_nc_ohc=CaoMpHba4cIQ7kNvgFGCanI\&_nc_zt=23\&_nc_ht=scontent-hkg4-1.xx\&_nc_gid=AN1f0kMjvYy10FgZlyKfKAX\&oh=00_AYC9q700UgP1h5gUiKeHCQDbziL6G2F4sd1eoXIVCPu8rQ\&oe=674E11E0}{SubFig.S7-7} (ANTA), SubFig.S7-8 (Apple), \\
& \href{https://img.alicdn.com/imgextra/i1/2215156393658/O1CN01PVqq501ctPlBVgLpe_!!2215156393658.jpg}{SubFig.S7-9} (Helen Keller) \\
&\\
\multirow{4}*{\textbf{Figure.~\ref{fig: user_s}}} & \href{https://www.zcool.com.cn/work/ZNzA0OTU2NjQ=.html}{SubFig.S8-1} (FIND VISUAL), \href{https://ifdesign.com/en/winner-ranking/project/nooie-robot-vacuum/580903}{SubFig.S8-2} (Nooie Robot Vacuum), \\ 
& \href{https://www.rimowa.com/on/demandware.static/-/Sites-rimowa-master-catalog-final/default/dw467d3040/images/large/92553004_2.png}{SubFig.S8-3} (RIMOWA), \href{https://manuals.plus/wp-content/uploads/2023/10/xiaomi-M2239B1-Smart-Band-8-Watch-Product-768x768.jpg?ezimgfmt=ng:webp/ngcb1}{SubFig.S8-4} (XIAOMI), \\
& \href{https://www.zcool.com.cn/work/ZNjY0NjAzNjA=.html}{SubFig.S8-5} (Box Studio), \href{https://images.meesho.com/images/products/433138062/mbgzu_512.webp}{SubFig.S8-6}, \\ 
& \href{https://www.zcool.com.cn/work/ZNjU0MTI2Njg=.html}{SubFig.S8-7} (Foodography), \href{https://www.zcool.com.cn/work/ZNjg5Njg0MjA=.html}{SubFig.S8-8} (FIND VISUAL) \\
&\\
\multirow{2}*{\textbf{Figure.~\ref{fig: qu_1}}} & \href{https://www.zcool.com.cn/work/ZNTI4NDEzODQ=.html}{SubFig.S9-1} (Xiangma), \href{https://www.zcool.com.cn/work/ZNTkyNTMzMDg=.html}{SubFig.S9-2} (METHOU), \\ 
& \href{https://www.zcool.com.cn/work/ZNDk2NzE3Mjg=.html}{SubFig.S9-3} (FILMSAYS), \href{https://images8.alphacoders.com/919/thumb-1920-919688.jpg}{SubFig.S9-4} (Alpha Coders) \\
&\\
\textbf{Figure.~\ref{fig: qu_2}} & \href{https://img.aupres.com.cn/aupres/product/main/2023122302.png?v=2023122701}{SubFig.S10-2} (AUPRES), \href{https://img.alicdn.com/imgextra/i3/2215050573719/O1CN01EsAOqi1dLLlGM2w86_!!2215050573719.jpg}{SubFig.S10-3}, \href{https://www.zcool.com.cn/work/ZNzA1NDU5NDQ=.html}{SubFig.S10-4} (MAOOXD) \\
&\\
\multirow{2}*{\textbf{Figure.~\ref{fig: qu_3}}} & \href{https://image.suning.cn/uimg/MZMS/show/166058363214958789.jpg}{SubFig.S11-1} (Bear), \href{https://gd-hbimg.huaban.com/c7a7100e82e3d1e6a254fa522399db9e611ffea19740b-ASz3ot_fw1200webp}{SubFig.S11-2} (Luckin Coffee), \\ 
& \href{https://store.storeimages.cdn-apple.com/8756/as-images.apple.com/is/airpods-pro-2-hero-select-202409_FV1?wid=976\&hei=916\&fmt=jpeg\&qlt=90\&.v=1725492499003}{SubFig.S11-3} (Apple), \href{https://www.zcool.com.cn/work/ZNjkxODQ0OTI=.html}{SubFig.S11-4} (FIND VISUAL) \\
&\\
\multirow{2}*{\textbf{Figure.~\ref{fig: gpt_4o}}} & SubFig.S12-1 (Bear),\href{https://www.zcool.com.cn/work/ZNjE1OTMxMjQ=.html}{SubFig.S12-2} (Foodography), \\
& \href{https://www.zcool.com.cn/work/ZNDM5NzEzMDg=.html}{SubFig.S12-3} (YUANYI). \\
\bottomrule
\end{tabular}
}
\end{table}

\begin{figure*}[t]
    \centering
    \includegraphics[width=0.95\textwidth]{sec/assets/vision_result.pdf}
    \caption{More cases generated by Pinco. Pinco supports the generation of high-quality images with different aspect ratios, while ensuring the reasonable placement of subjects, achieving realistic foreground-conditioned inpainting. The 11 images in the upper section are generated by HY-Pinco, while the 8 images in the lower section are produced by Flux-Pinco.}
    \label{fig: vs}
\end{figure*}

\begin{figure*}[t]
    \centering
    \includegraphics[width=0.85\textwidth]{sec/assets/RoPE_convergence.pdf}
    \caption{More cases of the model training process. We provide a detailed demonstration of the training process of the Pinco, while only presenting the final results of the Pinco-w/o RoPE due to limited space. We can observe that the Pinco model can gradually place the given subject into the generated scene while maintaining the contour and foreground consistency. In contrast, although the same number of training epochs were used, the Pinco-w/o RoPE model can only learn partial subject information, resulting in distortions in both contour and foreground consistency. Zoom in to observe the details.}
    \label{fig: rope}
\end{figure*}

\begin{figure*}[t]
    \centering
    \includegraphics[width=0.95\textwidth]{sec/assets/User_Study_Vision.pdf}
    \caption{Some cases in our user study.}
    \label{fig: user_s}
\end{figure*}

\begin{figure*}[t]
    \centering
    \includegraphics[width=0.85\textwidth]{sec/assets/More_Vision_Result.pdf}
    \caption{More qualitative comparisons between our Pinco and the state-of-the-art methods.}
    \label{fig: qu_1}
\end{figure*}

\begin{figure*}[t]
    \centering
    \includegraphics[width=0.85\textwidth]{sec/assets/More_Vision_Result_2.pdf}
    \caption{More qualitative comparisons between our Pinco and the state-of-the-art methods.}
    \label{fig: qu_2}
\end{figure*}

\begin{figure*}[t]
    \centering
    \includegraphics[width=0.85\textwidth]{sec/assets/More_Vision_Result_3.pdf}
    \caption{More qualitative comparisons between our Pinco and the state-of-the-art methods.}
    \label{fig: qu_3}
\end{figure*}

\begin{figure*}[t]
    \centering
    \includegraphics[width=0.8\textwidth]{sec/assets/GPT4O.pdf}
    \caption{GPT-4o rationality analysis results.}
    \label{fig: gpt_4o}
\end{figure*}


\end{document}